\documentclass[10pt,twocolumn,letterpaper]{article}

\usepackage{cvpr}
\usepackage{times}
\usepackage{epsfig}
\usepackage{graphicx}
\usepackage{amsmath}
\usepackage{amssymb}

\usepackage{booktabs}       
\usepackage{color}
\definecolor{darkgreen}{rgb}{0,0.6,0.2}

\usepackage[pagebackref=true,breaklinks=true,letterpaper=true,colorlinks,bookmarks=false]{hyperref}

\cvprfinalcopy 

\def\cvprPaperID{3515} 

\ifcvprfinal\fi
\begin{document}

\title{Towards Multi-pose Guided Virtual Try-on Network}


\author{
Haoye Dong$^{1}$ ,~ Xiaodan Liang$^{1}$ ,~ Bochao Wang$^1$ ,~ Hanjiang Lai$^{1}$ ,~ Jia Zhu$^2$ ,~ Jian Yin$^{1}$\\
$^1$Sun Yat-sen University,  $^2$South China Normal University\\
{\tt\small \{donghy7@mail2, wangboch@mail2, laihanj3@mail, issjyin@mail\}.sysu.edu.cn}, \\
{\tt\small xdliang328@gmail.com, jzhu@m.scun.edu.cn}
}

\maketitle

\begin{abstract}
   Virtual try-on system under arbitrary human poses has huge application potential, yet raises quite a lot of challenges, e.g. self-occlusions, heavy misalignment among diverse poses, and diverse clothes textures. Existing methods aim at fitting new clothes into a person can only transfer clothes on the fixed human pose, but still show unsatisfactory performances which often fail to preserve the identity, lose the texture details, and decrease the diversity of poses. In this paper, we make the first attempt towards multi-pose guided virtual try-on system, which enables transfer clothes on a person image under diverse poses. Given an input person image, a desired clothes image, and a desired pose, the proposed Multi-pose Guided Virtual Try-on Network (MG-VTON) can generate a new person image after fitting the desired clothes into the input image and manipulating human poses. Our MG-VTON is constructed in three stages: 1) a desired human parsing map of the target image is synthesized to match both the desired pose and the desired clothes shape; 2) a deep Warping Generative Adversarial Network (Warp-GAN) warps the desired clothes appearance into the synthesized human parsing map and alleviates the misalignment problem between the input human pose and desired human pose; 3) a refinement render utilizing multi-pose composition masks recovers the texture details of clothes and removes some artifacts. Extensive experiments on well-known datasets and our newly collected largest virtual try-on benchmark demonstrate that our MG-VTON significantly outperforms all state-of-the-art methods both qualitatively and quantitatively with promising multi-pose virtual try-on performances.
\end{abstract}

\section{Introduction}
\begin{figure}[!tp]
\centering
\includegraphics[width=1.0\hsize \hspace{0.01\hsize}]{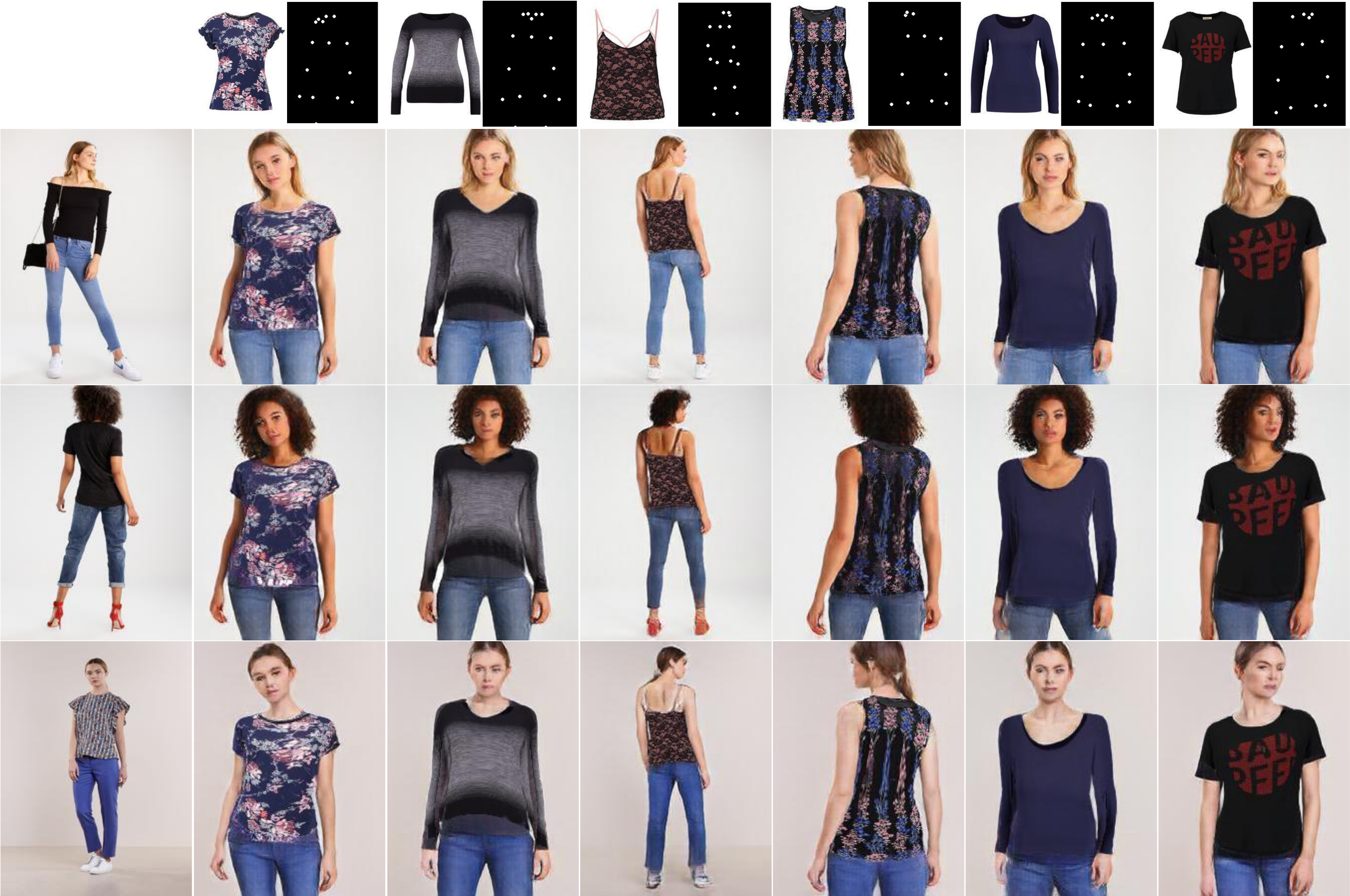} 
\caption{Some results of our model by manipulating both various clothes and diverse poses. The input image of the clothes and poses are shown in the first row, while the input images of the person are shown in the first column. The results manipulated by both clothes and pose are shown in the other columns.}
\label{fig:fig1}
\vspace{-6mm}
\end{figure}

Learning to synthesize the image of person conditioned on the image of clothes and manipulate the pose simultaneously is a significant and valuable task in many applications such as virtual try-on, virtual reality, and human-computer interaction. In this work, we propose a multi-stage method to synthesize the image of person conditioned on both clothes and pose. Given an image of a person, a desired clothes, and a desired pose, we generate the realistic image that preserves the appearance of both desired clothes and person, meanwhile reconstructing the pose, as illustrated in Figure~\ref{fig:fig1}. Obviously, delicate and reasonable synthesized outfit with arbitrary pose is helpful for users in selecting clothes while shopping.

However, recent image synthesis approaches~\cite{han2017viton,wang2018cpvton} for virtual try-on mainly focus on the fixed pose and fail to preserve the fine details, such as the clothing of lower-body and the hair of the person lose the details and style, as shown in Figure~\ref{fig:vs_others}. In order to generate the realistic image, those methods apply a coarse-to-fine network to produce the image conditioned on clothes only. They ignore the significant features of the human parsing, which leads to synthesize blurry and unreasonable image, especially in case of conditioned on various poses. For instance,  as shown in Figure~\ref{fig:vs_others}, the clothing of lower-body cannot be preserved while the clothing of upper-body is replaced. The head of the person fail to identify while conditioned different poses. Other exiting works~\cite{laehner2018deepwrinkles,pons2017clothcap,zhang2017detailed} usually leverage 3D measurements to solve those issues since the 3D information have abundant details of the shape of the body that can help to generate the realistic results. However, it needs expert knowledge and huge labor cost to build the 3D models, which requires collecting the 3D annotated data and massive computation. These costs and complexity would limit the applications in the practical virtual try-on simulation.

In this paper, we study the problem of virtual try-on conditioned on 2D images and arbitrary poses, which aims to learn a mapping function from an input image of a person to another image of the same person with a new outfit and diverse pose, by manipulating the target clothes and pose. Although the image-based virtual try-on with the fixed pose has been studied widely~\cite{han2017viton,wang2018cpvton,zhu2017fashionGAN}, the task of multi-pose virtual try-on is less explored. In addition, without modeling the mapping of the intricate interplay among of the appearance, the clothes, and the pose, directly using the existing virtual try-on methods to synthesized image based on different poses often result in blurry and artifacts. 

Targeting on the problems mentioned above, we propose a novel Multi-pose Guided Virtual Try-on Network (MG-VTON) that can generate a new person image after fitting both desired clothes into the input image and manipulating human poses. Our MG-VTON is a multi-stage framework with generative adversarial learning. Concretely, we design a pose-clothes-guided human parsing network to estimate a plausible human parsing of the target image conditioned on the approximate shape of the body, the face mask, the hair mask, the desired clothes, and the target pose, which could guide the synthesis in an effective way with the precise region of body parts. To seamlessly fit the desired clothes on the person, we warp the desired clothes image, by exploiting a geometric matching model to estimate the transformation parameters between the mask of the input clothes image and the mask of the synthesized clothes extracted from the synthesized human parsing. In addition, we design a deep Warping Generative Adversarial Network (Warp-GAN) to synthesize the coarse result alleviating the large misalignment caused by the different poses and the diversity of clothes. Finally, we present a refinement network utilizing multi-pose composition masks to recover the texture details and alleviate the artifact caused by the large misalignment between the reference pose and the target pose.

To demonstrate our model, we collected a new dataset, named MPV, by collecting various clothes image and person images with diverse poses from the same person. In addition, we also conduct experiments on DeepFashion~\cite{liu2016deepfashion} datasets for testing. Following the object evaluation protocol~\cite{wang2017pix2pixHD}, we conduct a human subjective study on the Amazon Mechanical Turk (AMT) platform. Both quantitative and qualitative results indicate that our method achieves effective performance and high-quality images with appealing details. The main contributions are listed as follows:

\begin{itemize}
\item A new task of virtual try-on conditioned on multi-pose is proposed, which aims to restructure the person image by manipulating both diverse poses and clothes.
\item We propose a novel Multi-pose Guided Virtual Try-on Network (MG-VTON) that generates a new person image after fitting the desired clothes onto the input person image and manipulating human poses. MG-VTON contains four modules: 1) a pose-clothes-guided human parsing network is designed to guide the image synthesis; 2) a Warp-GAN learns to synthesized realistic image by using a warping features strategy; 3) a refinement network learns to recover the texture details; 4) a mask-based geometric matching network is presented to warp clothes that enhances the visual quality of the generated image.
\item A new dataset for the multi-pose guided virtual try-on task is collected, which covers person images with more poses and clothes diversity. The extensive experiments demonstrate that our approach can achieve the competitive quantitative and qualitative results.
\end{itemize}

\begin{figure*}[!ht]
\centering
\includegraphics[width=0.9\hsize \hspace{0.01\hsize}]{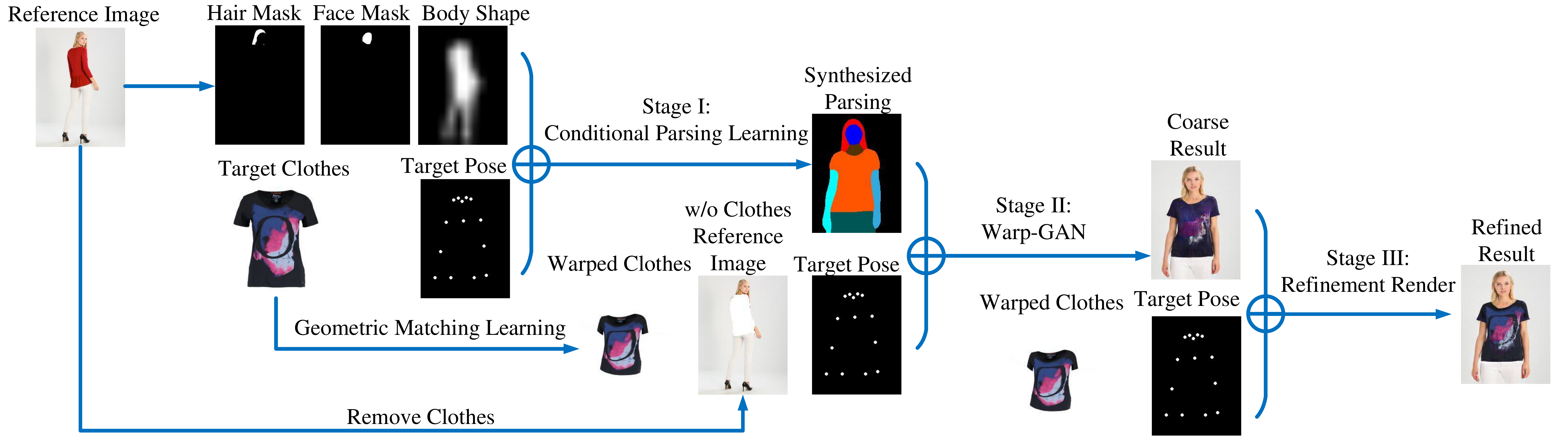} 
\caption{The overview of the proposed MG-VTON. Stage I: We first decompose the reference image into three binary masks. Then, we concatenate them with the target clothes and target pose as an input of the conditional parsing network to predict human parsing map. Stage II: Next, we warp clothes, remove the clothing from the reference image, and concatenate them with the target pose and synthesized parsing to synthesize the coarse result by using Warp-GAN. Stage III: We finally refine the coarse result with a refinement render, conditioning on the warped clothes, target pose, and the coarse result.}
\label{fig:test_pipeline}
\vspace{-4mm}
\end{figure*}

\section{Related Work}
\textbf{Generative Adversarial Networks (GANs).} 
GANs~\cite{goodfellow2014generative} consists of a generator and a discriminator that the discriminator learns to classify between the synthesized images and the real images while the generator tries to fool the discriminator. The generator aims to generate realistic images, which are indistinguishable from the real images. And the discriminator focuses on distinguishing between the synthesized and real images. Existing works have leveraged various applications based on GANs, such as style transfer~\cite{isola2017pix2pix,zhu2017cycleGAN,kim2017discoGAN,yi2017dualgan}, image inpainting~\cite{Yang2017inpainting}, text-to-image~\cite{reed2016text2image}, and super-resolution imaging~\cite{ledig2016photo}. Inspired by those impressive results of GANs, we also apply the adversarial loss to exploit a virtual try-on method with GANs.

\textbf{Person image synthesis.}
Skeleton-aided~\cite{yan2017skeleton} proposed a skeleton-guided person image generation method, which conditioned on a person image and the target skeletons. PG2~\cite{ma2017pose} applied a coarse-to-fine framework that consists of a coarse stage and a refined stage. Besides, they proposed a novel model~\cite{ma2017disentangled} to further improve the quality of result by using a decomposition strategy. The deformableGANs~\cite{siarohin2017deformable} and \cite{balakrishnan2018synthesizing} made attempt to alleviate the misalignment problem between different poses by using affine transformation on the coarse rectangle region and warped the parts on pixel-level, respectively. V-UNET~\cite{Esser2018vunet} introduced a variational U-Net~\cite{ronn2015unet} to synthesize the person image by restructuring the shape with stickman label. \cite{pumarola2018unsupervised} applied CycleGAN~\cite{zhu2017cycleGAN} directly to manipulate pose. However, all those works fail to preserve the texture details consistency corresponding with the pose. The reason behind that is they ignore to consider the interplay between the human parsing map and the pose in the person image synthesis. The human parsing map can guide the generator to synthesize image in the precise region level that ensures the coherence of body structure.

\textbf{Virtual try-on.}
VITON~\cite{han2017viton} and  CP-VTON~\cite{wang2018cpvton} all presented an image-based virtual try-on network, which can transfer a desired clothes on the person by using a warping strategy. VITON computed the transformation mapping by the shape context TPS warps~\cite{belongie2002shape} directly. CP-VTON introduced a learning method to estimate the transformation parameters. FashionGAN~\cite{zhu2017fashionGAN} learned to generate new clothes on the input image of the person conditioned on a sentence describing the different outfit. However, the above all methods synthesized the image of person only on the fixed pose, which limits the applications in the practical virtual try-on simulation. ClothNet~\cite{lassner2017generative} presented an image-based generative model to produce new clothes conditioned on color. CAGAN~\cite{jetchev2017conditional} proposed a conditional analogy network to synthesize person image conditioned on the paired of clothes, which limits the practical virtual try-on scenarios. In order to generate the realistic-look person image in different clothes, ClothCap~\cite{pons2017clothcap} utilized the 3D scanner to capture the clothes, the shape of the body automatically. \cite{sekine2014virtual} presented a virtual fitting system that requires the 3D body shape, which is laborious for collecting the annotation. In this paper, we introduce a novel and effective method for learning to synthesize image with the new outfit on the person through adversarial learning, which can manipulate the pose simultaneously.

\begin{figure*}[!ht]
\centering
\includegraphics[width=1.0\hsize \hspace{0.01\hsize}]{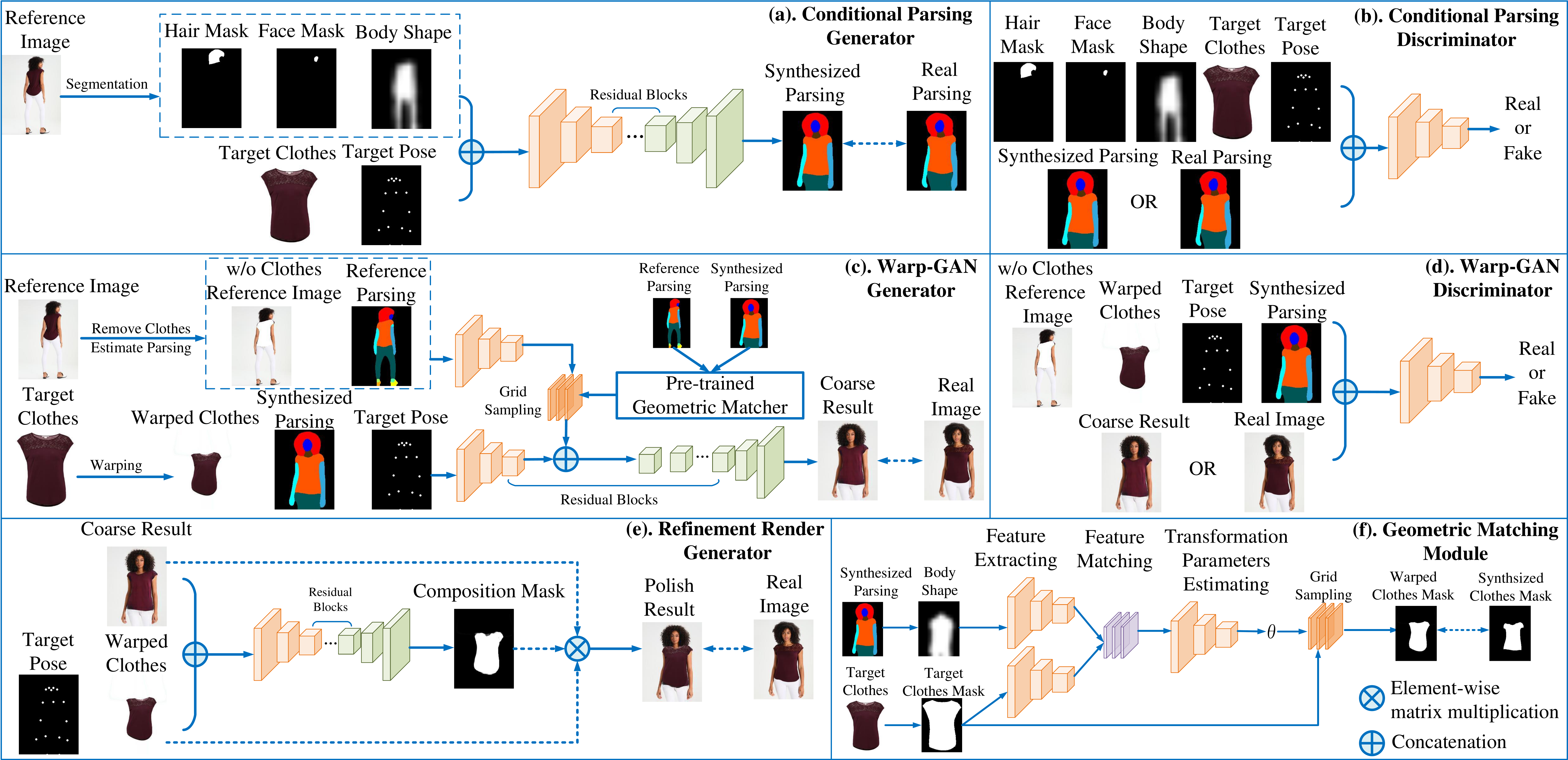} 
\caption{The network architecture of the proposed MG-VTON. (a)(b): The conditional parsing learning module consists of a pose-clothes-guided network that predicts the human parsing, which helps to generate high-quality person image. (c)(d): The Warp-GAN learns to generate the realistic image by using a warping features strategy due to the misalignment caused by the diversity of pose. (e): The refinement render network learns the pose-guided composition mask that enhances the visual quality of the synthesized image. (f): The geometric matching network learns to estimate the transformation mapping conditioned on the body shape and clothes mask.}
\label{fig:train_pipeline}
\vspace{-4mm}
\end{figure*}

\section{MG-VTON}
We propose a novel Multi-pose Guided Virtual Try-on Network (MG-VTON) that learns to synthesize the new person image for virtual try-on by manipulating both clothes and pose. Given an input person image, a desired clothes, and a desired pose, the proposed MG-VTON aims to produce a new image of the person wearing the desired clothes and manipulating the pose. Inspired by the coarse-to-fine idea~\cite{han2017viton,ma2017pose}, we adopt an outline-coarse-fine strategy that divides this task into three subtasks, including the conditional parsing learning, the Warp-GAN, and the refinement render. The Figure~\ref{fig:test_pipeline} illustrates the overview of MG-VTON. 

We first apply the pose estimator~\cite{cao2017openpose} to estimate the pose. Then, we encode the pose as 18 heatmaps, which is filled with ones in a circle with radius 4 pixels and zeros elsewhere. A human parser~\cite{gong2017look} is used to predict the human segmentation maps, consisting of 20 labels, from which we extract the binary mask of the face, the hair, and the shape of the body. Following VITON~\cite{han2017viton}, we downsample the shape of the body to a lower resolution ($16 \times 12$) and directly resize it to the original resolution  ($256 \times 192$), which alleviates the artifacts caused by the variety of the body shape.

\subsection{Conditional Parsing Learning}
To preserve the structural coherence of the person while manipulating both clothes and the pose, we design a pose-clothes-guided human parsing network, conditioned on the image of clothes, the pose heatmap, the approximated shape of the body, the mask of the face, and the mask of hair. As shown in Figure~\ref{fig:vs_others}, the baseline methods failed to preserve some parts of the person (e.g., the color of the trousers and the style of hair were replaced.), due to feeding the image of the person and clothes into the model directly. In this work, we leverage the human parsing maps to address those problems, which can help generator to synthesize the high-quality image on parts-level.

Formally, given an input image of person $I$, an input image of clothes $C$, and the target pose $P$, this stage learns to predict the human parsing map $S^{'}_t$ conditioned on clothes $C$ and the pose $P$. As shown in Figure~\ref{fig:train_pipeline} (a), we first extract the hair mask $M_h$, the face mask $M_f$, the body shape  $M_b$, and the target pose $P$ by using a human parser~\cite{gong2017look} and a pose estimator~\cite{cao2017openpose}, respectively. We then concatenate them with the image of clothes as input which is fed into the conditional parsing network. The inference of $S^{'}_t$ can be formulate as maximizing the posterior probability $p(S^{'}_t | (M_h, M_f, M_b, C, P))$. Furthermore, this stage is based on the conditional generative adversarial network (CGAN)~\cite{mirza2014cgan} which generates promising results on image manipulating. Thus, the poster probability $p(S^{'}_t | (M_h, M_f, M_b, C, P))$ is expressed as:
\begin{equation}
p(S^{'}_t | (M_h, M_f, M_b, C, P)) = G(M_h, M_f, M_b, C, P).
\end{equation}
We adopt a ResNet-like network as the generator $G$ to build the conditional parsing model. We adopt the discriminator $D$ directly from the pix2pixHD~\cite{wang2017pix2pixHD}. We apply the L1 loss for further improving the performance, which is advantageous for generating more smooth results~\cite{yan2017skeleton}. Inspired by the LIP~\cite{gong2017look}, we apply the pixel-wise softmax loss to encourage the generator to synthesize high-quality human parsing maps. Therefore, the problem of conditional parsing learning can be formulated as:
\begin{equation}
\begin{aligned}
    &\min_{G} \max_{D} F(G, D) \\
    & = \mathbb{E}_{M, C, P \sim p_{\text{data}}}[ \log (1 - D(G(M, C, P), M, C, P))]  \\
    & + \mathbb{E}_{S_t, M, C, P \sim p_{\text{data}}}[\log D(S_t, M, C, P)] \\
    & + \mathbb{E}_{S_t, M, C, P \sim p_{\text{data}}}[\| S_t - G(M, C, P)\|_1] \\ 
    & + \mathbb{E}_{S_t, M, C, P \sim p_{\text{data}}}[\mathcal{L}_{\text{parsing}}(S_t, G(M, C, P))],
   \label{eq:parsing}
\end{aligned}
\end{equation}
where $M$ denotes the concatenation of $M_h, M_f$, and $M_b$. The loss $\mathcal{L}_{\text{parsing}}$ denotes the pixel-wise softmax loss~\cite{gong2017look}. The $S_t$ denotes the ground truth human parsing. The $p_{\text{data}}$ represents the distributions of the real data. 

\subsection{Warp-GAN}
\label{s:wg}
Since the misalignment of pixels would lead to generate the blurry results~\cite{siarohin2017deformable}, we introduce a deep Warping Generative Adversarial Network (Warp-GAN) warps the desired clothes appearance into the synthesized human parsing map, which alleviates the misalignment problem between the input human pose and desired human pose. Different from deformableGANs~\cite{siarohin2017deformable} and \cite{balakrishnan2018synthesizing}, we warp the feature map from the bottleneck layer by using both the affine and TPS (Thin-Plate Spline)~\cite{bookstein1989tps} transformation rather than process the pixel directly by using affine only. Thanks to the generalization capacity of~\cite{Rocco2017geocnn}, we directly use the pre-trained model of \cite{Rocco2017geocnn} to estimate the transformation mapping between the reference parsing and the synthesized parsing. We then warp the w/o clothes reference image by using this transformation mapping. 

As illustrated in Figure~\ref{fig:train_pipeline} (c) and (d), the proposed deep warping network consists of the Warp-GAN generator $G_{\text{warp}}$ and the Warp-GAN discriminator $D_{\text{warp}}$. We use the geometric matching module to warp clothes image, as described in the section~\ref{s:gml}. Formally, we take warped clothes image $C_{\text{w}}$, w/o clothes reference image $I_{\text{w/o\_clothes}}$, the target pose $P$, and the synthesized human parsing $S^{'}_t$ as input of the Warp-GAN generator and synthesize the result $\hat{I} = G_{\text{warp}}(C_{\text{w}}, I_{\text{w/o\_clothes}}, P, S^{'}_t)$.
Inspired by~\cite{johnson2016perceptual,han2017viton,ledig2016photo}, we apply a perceptual loss to measure the distances between high-level features in the pre-trained model, which encourages generator to synthesize high-quality and realistic-look images. We formulate the perceptual loss as:
\begin{equation}
	\mathcal{L}_{\text{perceptual}}(\hat{I}, I) = \sum_{i=0}^{n} \alpha_{i} \| \phi_{i}(\hat{I}) - \phi_{i}(I) \|_{1},
\end{equation}
where $\phi_{i}(I)$ denotes the $i$-th $(i=0,1,2,3,4)$ layer feature map in pre-trained network $\phi$ of ground truth image $I$. We use the pre-trained VGG19~\cite{simonyan2015very} as $\phi$ and weightedly sum the L1 norms of last five layer feature maps in $\phi$ to represent perceptual losses between images. The $\alpha_{i}$ controls the weight of loss for each layer. 
In addition, following pixp2pixHD~\cite{wang2017pix2pixHD}, due to the feature map at different scales from different layers of discriminator enhance the performance of image synthesis, we also introduce a feature loss and formulate it as: 
\begin{equation}
	\mathcal{L}_{\text{feature}}(\hat{I}, I) = \sum_{i=0}^{n} \gamma_{i} \| F_{i}(\hat{I}) - F_{i}(I) \|_{1},
    \label{eq:feature}
\end{equation}
where $F_i(I)$ represent the $i$-th $(i=0,1,2)$ layer feature map of the trained $D_{\text{warp}}$. The $\gamma_{i}$ denotes the weight of L1 loss for corresponding layer.

Furthermore, we also apply the adversarial loss $\mathcal{L}_{\text{adv}}$~\cite{goodfellow2014generative,mirza2014cgan} and L1 loss $\mathcal{L}_{\text{1}}$~\cite{yan2017skeleton} to improve the performance. We design a weight sum losses as the loss of $G_{\text{warp}}$, which encourages the $G_{\text{warp}}$ to synthesize realistic and natural images in different aspects, written as follows:
\begin{equation}
	\mathcal{L}_{G_{\text{warp}}} = \lambda_{1} \mathcal{L}_{\text{adv}}  +  \lambda_{2} \mathcal{L}_{\text{perceptual}} +  \lambda_{3} \mathcal{L}_{\text{feature}} +  \lambda_{4} \mathcal{L}_{\text{1}},
    \label{eq:loss}
\end{equation}
 where $\lambda_{i}$ $(i=1,2,3,4)$ denotes the weight of corresponding loss, respectively. 

\subsection{Refinement render}
\label{s:rr}
In the coarse stage, the identification information and the shape of the person can be preserve, but the texture details are lost due to the complexity of the clothes image. Pasting the warped clothes onto the target person directly may lead to generate the artifacts. Learning the composition mask between the warped clothes image and the coarse results also generates the artifacts~\cite{han2017viton,wang2018cpvton} due to the diversity of pose. To solve the above issues, we present a refinement render utilizing multi-pose composition masks to recover the texture
details and remove some artifacts.

Formally, we define $C_{\text{w}}$ as an image of warped clothes obtained by geometric matching learning module, $\hat{I_{\text{c}}}$ as a coarse result generated by the Warp-GAN, $P$ as the target pose heatmap, and $G_{\text{p}}$ as the generator of the refinement render. As illustrated in Figure~\ref{fig:train_pipeline} (e), taking $C_{\text{w}}$, $\hat{I_{\text{c}}}$, and $P$ as input, the $G_{\text{p}}$ learns to predict a towards multi-pose composition mask and synthesize the rendered result:
\begin{equation}
\hat{I}_{\text{p}} = G_{\text{p}}(C_{\text{w}}, \hat{I}, P) \odot C_{\text{w}} + (1 - G_{\text{p}}(C_{\text{w}}, \hat{I}, P)) \odot \hat{I},
\end{equation}
where $\odot$ denotes the element-wise matrix multiplication. We also adopt the perceptual loss to enhance the performance that the objective function of $G_{\text{p}}$ can be written as:
\begin{equation}
\mathcal{L}_{\text{p}} = \mu_{1} \mathcal{L}_{\text{perceptual}}(\hat{I}_{\text{p}}, I) + \mu_{2} \| 1 - G_{\text{p}}(C_{\text{w}}, \hat{I_{\text{c}}}, P)  \|_{1},
\end{equation}
where $\mu_{1}$ denotes the weight of perceptual loss and $\mu_{2}$ denotes the weight of the mask loss.

\subsection{Geometric matching learning}
\label{s:gml}
Inspired by~\cite{Rocco2017geocnn}, we adopt the convolutional neural network to learn the transformation parameters, including feature extracting layers, feature matching layers, and the transformation parameters estimating layers. As shown in Figure~\ref{fig:train_pipeline} (f), we take the mask of the clothes image, and the mask of body shape as input which is first passed through the feature extracting layers. Then, we predict the correlation map by using the matching layers. Finally, we apply a regression network to estimate the TPS (Thin-Plate Spline)~\cite{bookstein1989tps} transformation parameters for the clothes image directly based on the correlation map. 

Formally, given an input image of clothes $C$ and its mask $C_{mask}$, following the stage of conditional parsing learning, we obtain the approximated body shape $M_b$ and the synthesized clothes mask $\hat{C}_{mask}$ from the synthesized human parsing. This subtask aims to learn the transformation mapping function $\mathcal{T}$ with parameter $\theta$ for warping the input image of clothes $C$. Due to the unseen of synthesized clothes but have the synthesized clothes mask, we learn the mapping between the original clothes mask $C_{mask}$ and the synthesized clothes mask $\hat{C}_{mask}$ obey body shape $M_b$. Thus, the objective function of the geometric matching learning can be formulated as:
\begin{equation}
\mathcal{L}_{\text{geo\_matching}}(\theta) = \| \mathcal{T}_{\theta}(C_{mask}) -  \hat{C}_{mask}\|_1,
\label{eq:matching}
\end{equation}
Therefore, the warped clothes $C_{w}$ can be formulated as $C_{w} = \mathcal{T}_{\theta}(C)$, which is helpful for addressing the problem of misalignment and learning the composition mask in the above subsection~\ref{s:wg} and subsection~\ref{s:rr}.

\begin{figure*}[!tp]
\centering
\includegraphics[width=.8\hsize \hspace{0.01\hsize}]{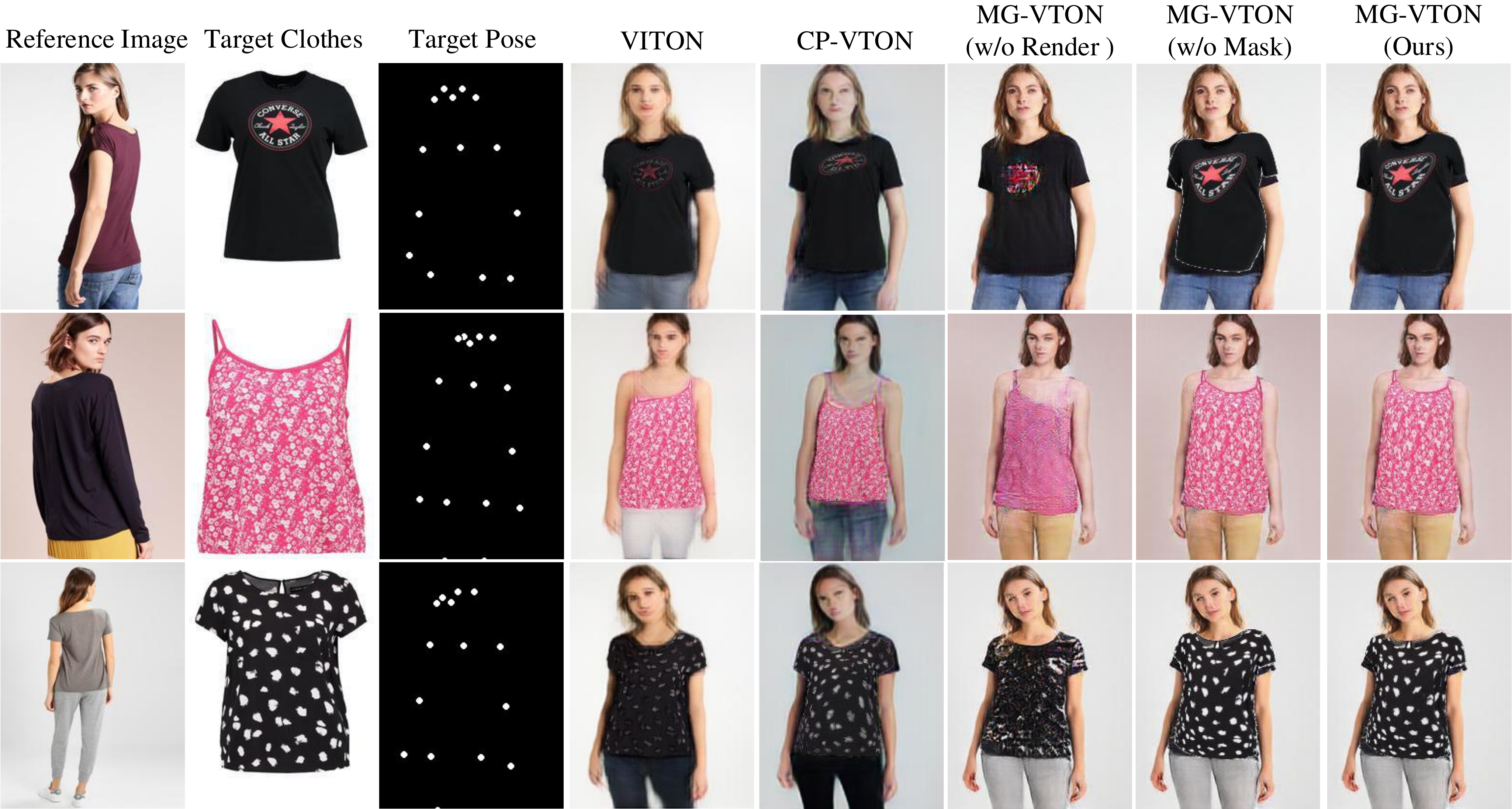} 
\caption{Visualized comparison with other methods on our collected dataset MPV. MG-VTON (w/o Render) is the model where the refinement render is removed. The model where the multi-pose composition mask is removed denotes as MG-VTON (w/o Mask).}
\label{fig:vs_others}
\vspace{-4mm}
\end{figure*}

\section{Experiments}
In this section, we first make visual comparisons with other methods and then discuss the results quantitatively. We also conduct the human perceptual study and the ablation study, and further train our model on our newly collected dataset MPV test it on the Deepfashion to verify the generation capacity.

\subsection{Datasets}
Since each person image in the dataset used in VITON~\cite{han2017viton} and CP-VTON~\cite{wang2018cpvton} only has one fixed pose, we collected the new dataset from the internet, named MPV, which contains 35,687 person images and 13,524 clothes images. Each person image in MPV has different poses. The image is in the resolution of $256 \times 192$. We extract the 62,780 three-tuples of the same person in the same clothes but with diverse poses. We further divide them into the train set and the test set with 52,236 and 10,544 three-tuples, respectively. Note that we shuffle the test set with different clothes and diverse pose for quality evaluation. DeepFashion~\cite{liu2016deepfashion} only have the pairs of the same person in different poses but do not have the image of clothes. To verify the generalization capacity of the proposed model, we extract 10,000 pairs from DeepFashion, and randomly select clothes image from the test set of the MPV for testing.

\subsection{Evaluation Metrics}
We apply three measures to evaluate the proposed model, including subjective and objective metrics:
1) We perform pairwise A/B tests deployed on the Amazon Mechanical Turk (\textbf{AMT}) platform for human perceptual study. 2) we use Structural SIMilarity (\textbf{SSIM})~\cite{wang2004image} to measure the similarity between the synthesized image and ground truth image. In this work, we take the target image (the same person wearing the same clothes) as the ground truth image used to compare with the synthesized image for computing SSIM. 3) We use Inception Score (\textbf{IS})~\cite{NIPS2016_6125} to measure the quality of the generated images, which is a common method to verify the performances for image generation.
 
\subsection{Implementation Details}
\textbf{Setting.}
We train the conditional parsing network, Warp-GAN, refinement render, and geometric matching network for 200, 15, 5, 35 epochs, respectively, using ADAM optimizer~\cite{Diederik2014Adam}, with the batch size of 40, learning rate of 0.0002, $\beta_1 = 0.5$, $\beta_2 = 0.999$. We use two NVIDIA Titan XP GPUs and Pytorch platform on Ubuntu 14.04.

\textbf{Architecture.}
As shown in Figure~\ref{fig:train_pipeline}, each generator of MG-VTON is a ResNet-like network, which consists of three downsample layers, three upsample layers, and nine residual blocks, each block has three convolutional layers with 3x3 filter kernels followed by the bath-norm layer and Relu activation function. Their number of filters are 64, 128, 256, 512, 512, 512, 512, 512, 512, 512, 512, 512, 256, 128, 64. For the discriminator, we apply the same architecture as pix2pixHD~\cite{wang2017pix2pixHD}, which can handle the feature map in different scale with different layers. Each discriminator contains four downsample layers which include 4x4 kernels, InstanceNorm, and LeakyReLU activation function. 

\subsection{Baselines}
\textbf{VITON}~\cite{han2017viton} and \textbf{CP-VTON}~\cite{wang2018cpvton} are the state-of-the-art image-based virtual try-on method which assumes the pose of the person is fixed. They all used warped clothes image to improve the visual quality, but lack of the ability to generate image under arbitrary poses. In particular, VTION directly applied shape context matching~\cite{belongie2002shape} to compute the transformation mapping. CP-VTON borrowed the idea from \cite{Rocco2017geocnn} to estimate the transformation mapping using a convolutional network. To obtain fairness, we first enriched the input of the VITON and CP-VTON by adding the target pose. Then, we retrained the VITON and CP-VTON on MPV dataset with the same splits (train set and test set) as our model.

\subsection{Quantitative Results}
We conduct experiments on two benchmarks and compare against two recent related works using two widely used metrics \textbf{SSIM} and \textbf{IS} to verify the performance of the image synthesis, summarized in Table.~\ref{tab:ssim_is}, higher scores are better. The results shows that ours proposed methods significantly achieve higher scores and consistently outperform all baselines on both datasets thanks to the cooperation of our conditional parsing generator, Warp-GAN, and the refinement render. Note that the MG-VTON (w/o Render) achieves the best SSIM score and the MG-VTON (w/o Mask) achieve the best IS score, but they obtain worse visual quality results and achieve lower scores in AMT study compare with MG-VTON (ours), as illustrated in the Table~\ref{tab:amt} and Figure~\ref{fig:ab_polish}. As shown in Figure~\ref{fig:vs_others}, MG-VTON (ours) synthesizes more realistic-looking results than MG-VTON (w/o Render), but the latter achieve higher SSIM score, which also can be observed in~\cite{johnson2016perceptual}. Hence, we believe that the proposed MG-VTON can generate high-quality person image for multi-pose virtural try-on with convincing results.

\begin{table}[htbp]
\centering
\caption{Comparisons on MPV and DeepFashion. }
\vspace{2mm}
\resizebox{\columnwidth}{!}{
\begin{tabular}{lccc}
	\toprule
        & \multicolumn{2}{c}{MPV} &  \multicolumn{1}{c}{DeepFashion} \\
     \cmidrule{2-4} 
        Model & SSIM & IS   & IS  \\
    \midrule
        VITON~\cite{han2017viton}     		& 0.639    & 2.394 $\pm$ 0.205    & 2.302 $\pm$ 0.116 \\
        CP-VTON~\cite{wang2018cpvton}      	& 0.705	    & 2.519 $\pm$ 0.107     & 2.459 $\pm$ 0.212	\\ 
        MG-VTON (w/o Render)      	& \textbf{0.754}   &  2.694 $\pm$ 0.119   &	2.813 $\pm$ 0.047 \\ 
        MG-VTON (w/o Mask)      	& 0.733	   & \textbf{3.309 $\pm$ 0.137}    & \textbf{3.368 $\pm$ 0.055}	\\ 
        MG-VTON (Ours)		& 0.744	& 3.154 $\pm$ 0.142  &  3.030 $\pm$ 0.057	\\
    \bottomrule
\end{tabular}
}
\label{tab:ssim_is}
\vspace{-2mm}
\end{table}

\subsection{Qualitative Results}
We perform visual comparisons of the proposed method with VITON~\cite{han2017viton}, CP-VTON~\cite{wang2018cpvton}, MG-VTON (w/o Render), and MG-VTON (w/o Mask), illustrated in Figure~\ref{fig:vs_others}, which shows that our model generates reasonable results with convincing details. Although the baseline methods have synthesized few details of clothes, it is far from the practice towards multi-pose virtual try-on scenario. Specifically, the identity and the clothing of the lower-body cannot be preserved by the baseline methods. Besides, the clothing of the lower-body also cannot be preserved while the clothing of upper-body is change by the baseline methods. Furthermore, the baseline methods cannot synthesize the hairstyle and face well that result in blurry and artifacts. The reasons behind are that they overlook the high-level semantics of the reference image and the relationship between the reference image and target pose in the virtual try-on task. On the contrary, we adopt clothes and pose guided network to generate the target human parsing, which is helpful to alleviate the problem that lower-body clothing and hair cannot be preserved. In addition, we also design a deep warping network with an adversarial loss carefully to solve the issue that the identity cannot be preserved. Furthermore, we capture the interplay of among the poses and present a multi-pose based refined network that learns to erase the noise and artifacts. 

\begin{figure}[!hp]
\centering
\includegraphics[width=1.0\hsize \hspace{0.01\hsize}]{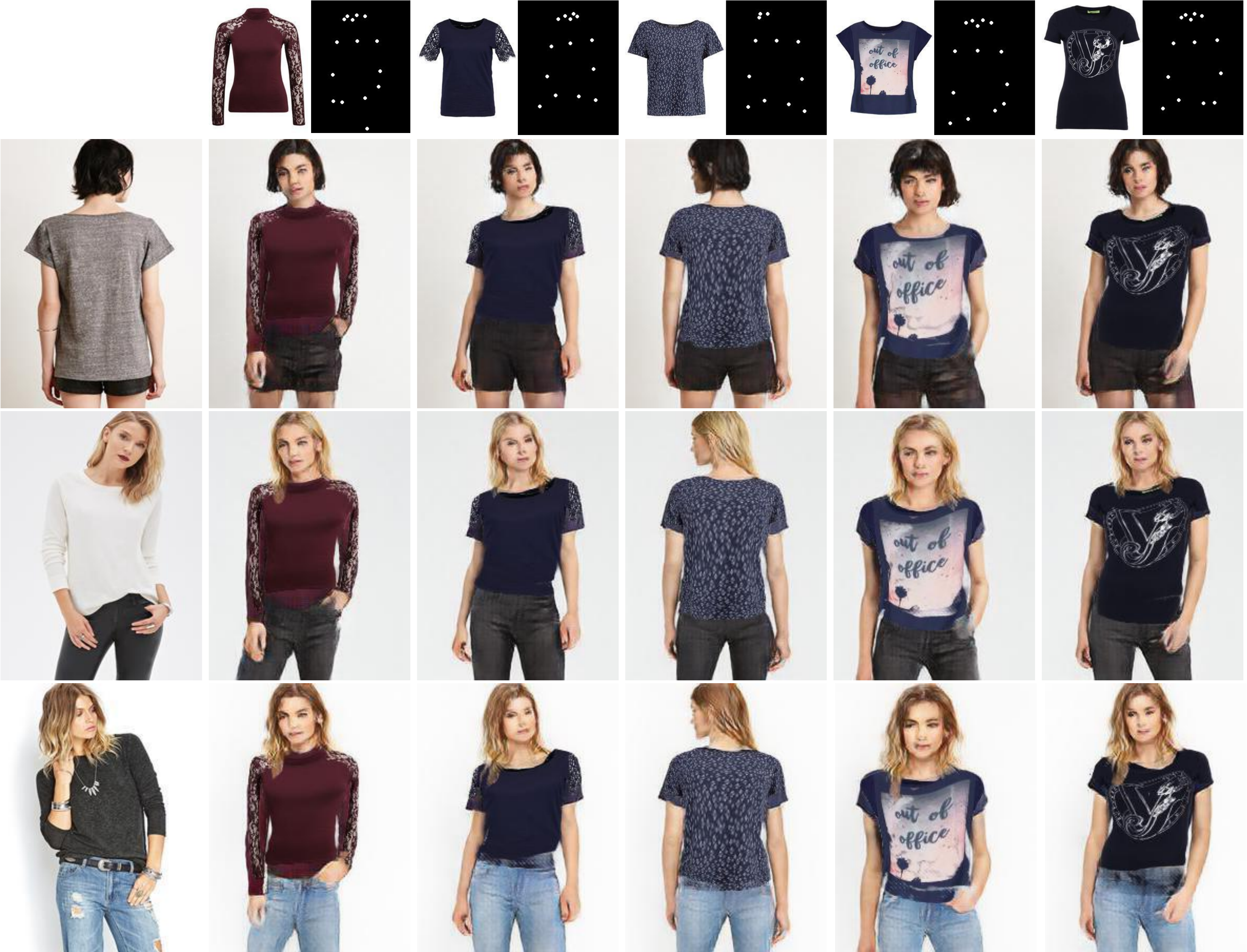} 
\caption{Some results from our model trained on MPV and tested on DeepFashion, which synthesizes the realistic image and captures the desired pose and clothes well.}
\label{fig:teston_dp}
\vspace{-4mm}
\end{figure}

\begin{table}[h]
\centering
\caption{Pairwise comparison on MPV and DeepFashion. Each cell lists the percentage where our MG-VTON is preferred over the other method. Chance is at 50\%. }
\vspace{2mm}
\resizebox{\columnwidth}{!}{
\begin{tabular}{ccccc}
    \toprule
                 & VITON & CP-VTON &  MG-VTON  & MG-VTON   \\
                  &  &  &  (w/o Render) & (w/o Mask)   \\
    \midrule
         MPV       &    83.1\%    & 85.9\%  & 82.4\%  & 84.6\%   \\
         DeepFashion      &    88.9\%    & 83.3\%  & 84.6\% & 75.5\%      \\ 
    \bottomrule
\end{tabular}
}
\label{tab:amt}
\vspace{-4mm}
\end{table}
\subsection{Human Perceptual Study}
We perform a human study on MPV and Deepfashion~\cite{liu2016deepfashion} to evaluate the visual quality of the generated image. Similar to pix2pixHD~\cite{wang2017pix2pixHD}, we deployed the A/B tests on the Amazon Mechanical Turk (AMT) platform. There are 1,600 images with size $256 \times 192$. We have shown three images for reference (reference image, clothes, pose) and two synthesized images with the option for picking. The workers are given two choices with unlimited time to pick the one image looks more realistic and natural, considering how well target clothes and pose are captured and whether the identity and the appearance of the person are preserved. Specifically, the workers are shown the reference image, target clothes, target pose, and the shuffled image pairs. We collected 8,000 comparisons from 100 unique workers. As illustrated in Table~\ref{tab:amt}, the image synthesized by our model obtained higher human evaluation scores and indicate the high-quality results compare to the baseline methods.

\begin{figure*}[!thp]
\centering
\includegraphics[width=1.0\hsize \hspace{0.01\hsize}]{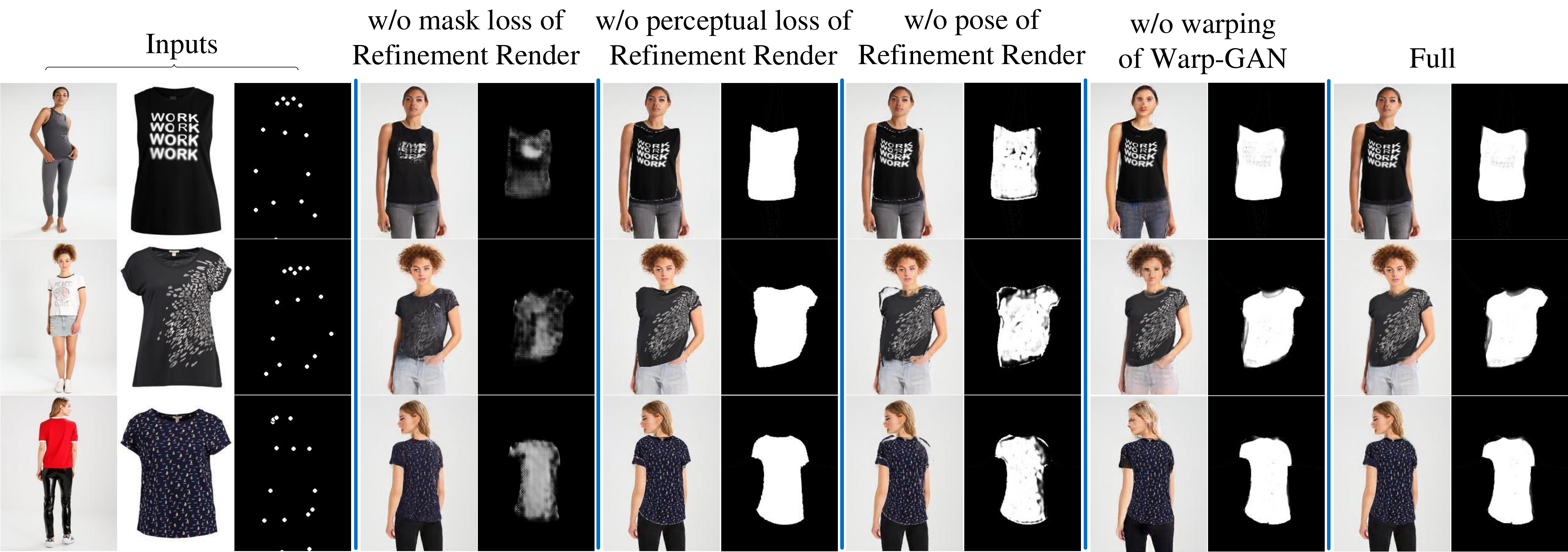} 
\caption{Ablation study on our collected dataset MPV. Zoom in for details.}
\label{fig:ab_polish}
\vspace{-4mm}
\end{figure*}

\begin{figure}[!tp]
\centering
\includegraphics[width=1.0\hsize \hspace{0.01\hsize}]{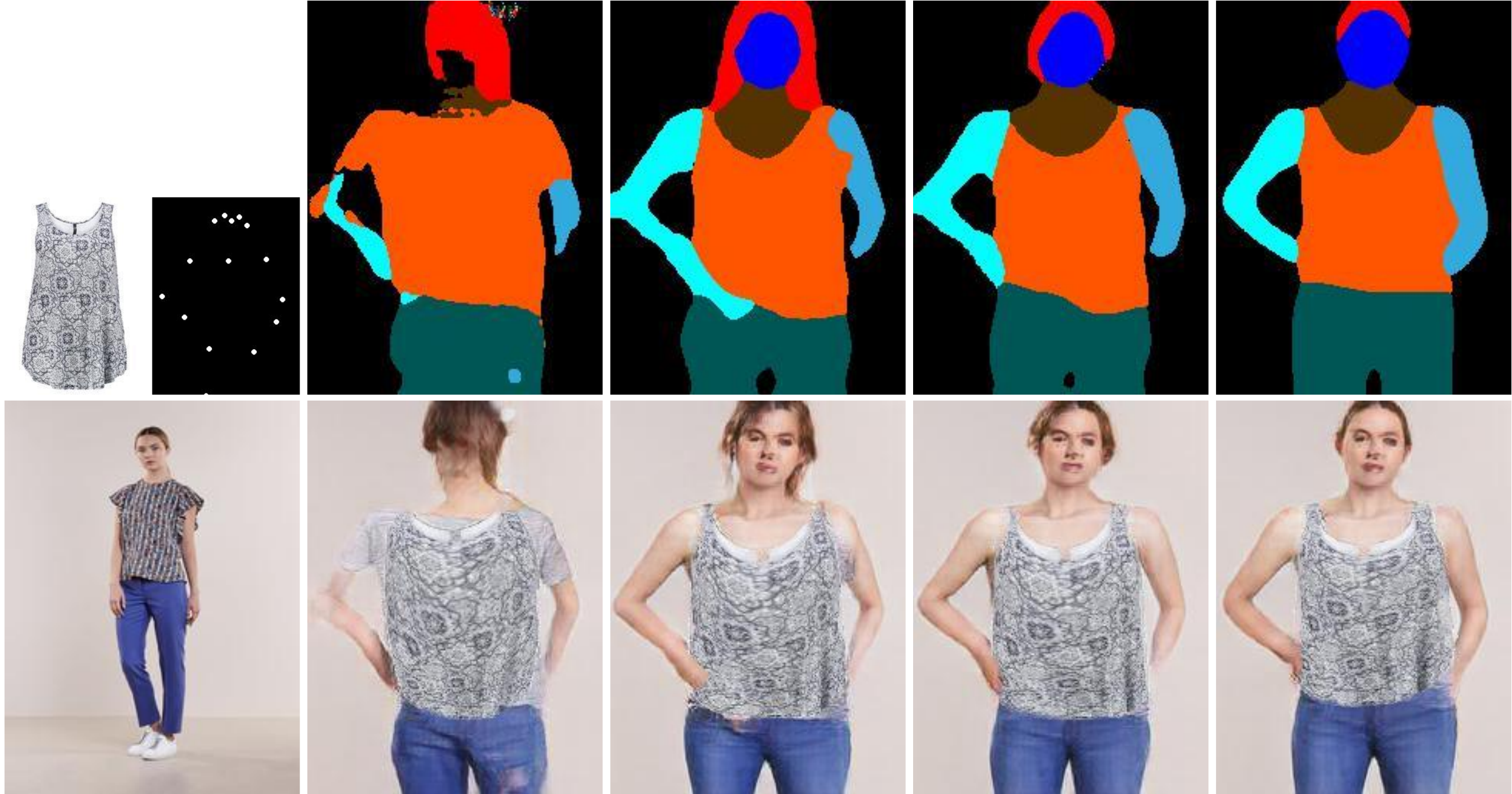} 
\caption{Effect of the quality of human parsing. The quality of human parsing significantly affects the quality of the synthesized image in the virtual try-on task.}
\label{fig:diff_parsing}
\vspace{-4mm}
\end{figure}

\subsection{Ablation Study}
We conduct an ablation study to analyze the important parts of our method. Observed from Table.~\ref{tab:ssim_is}, MG-VTON (w/o Mask) achieves the best scores. However, as shown in Figure~\ref{fig:vs_others}, it may inevitably generate artifacts. In Figure~\ref{fig:ab_polish}, we further evaluate the effect of the components of our MG-VTON. It shows that the multi-pose composition mask loss, the perceptual loss, and the pose in the refinement render stage, and the warping module in Warp-GAN are all important to enhance the performance.

\begin{figure}[!tp]
\centering
\includegraphics[width=1.0\hsize \hspace{0.01\hsize}]{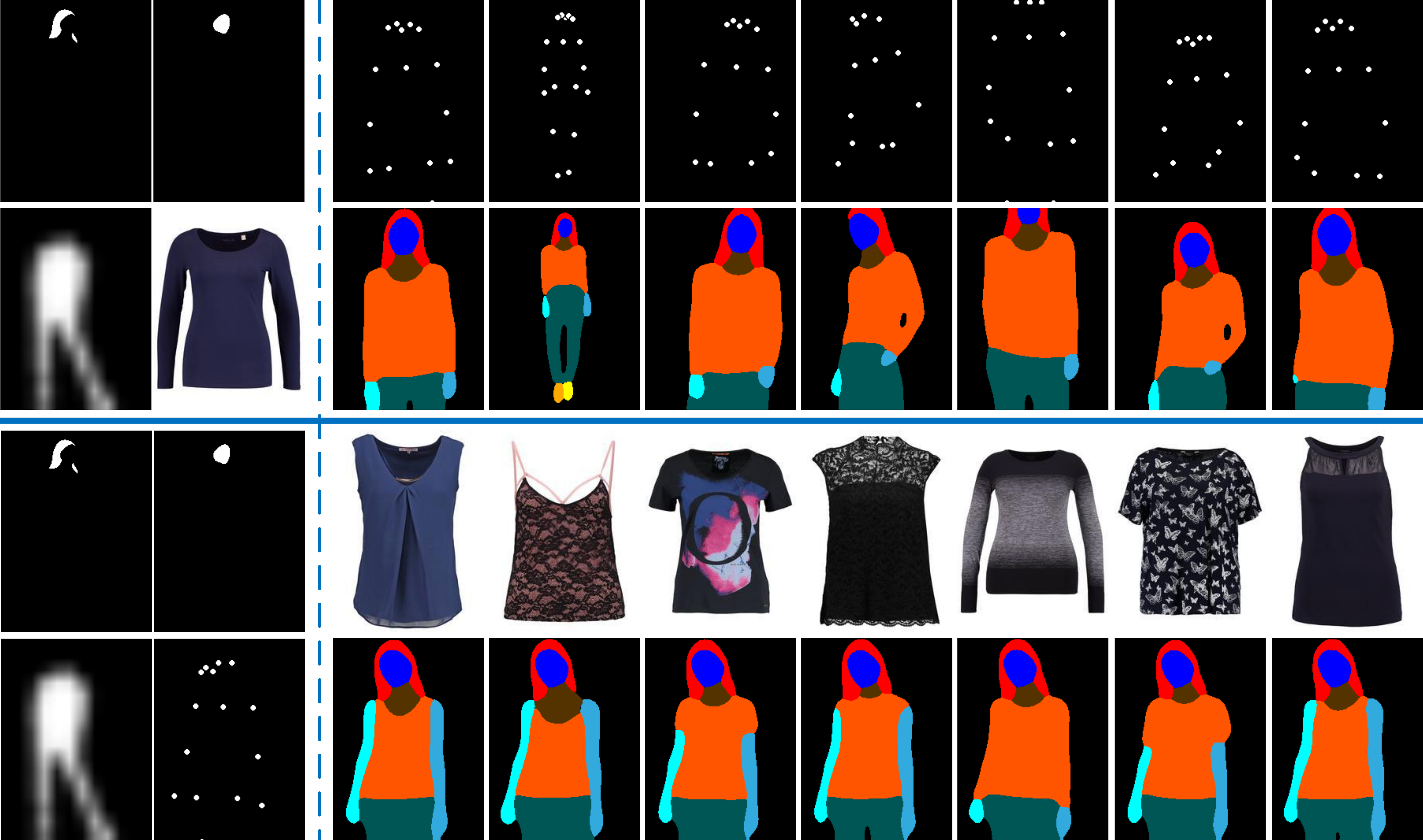} 
\caption{Effect of clothes and pose for the human parsing, which is manipulating by the pose and the clothes. }
\label{fig:diff_pose}
\vspace{-4mm}
\end{figure}

We also conduct an experiment to verify the effect of the human parsing in our MG-VTON. As shown in Figure~\ref{fig:diff_parsing}, there is a positive correlation between the quality of the human parsing with that of the result. We further to verify the effect of the synthesized human parsing by manipulating the desired pose and clothes, as illustrated in Figure~\ref{fig:diff_pose}. 
We manipulate the human parsing instead of the person image directly, and we can synthesize the person image in an easier and more effective way. Furthermore, we introduce an experiment that trained on our collected dataset MPV and test on the DeepFashion dataset to verify the generalization of the proposed model. As the Figure~\ref{fig:teston_dp} shown, our model captures the target pose and clothes well.

\section{Conclusions}
In this work, we make the first attempt to investigate the multi-pose guided virtual try-on system, which enables clothes transferred onto a person image under diverse poses. We propose a Multi-pose Guided Virtual Try-on Network (MG-VTON) that generates a new person image after fitting the desired clothes into the input image and manipulating human poses. Our MG-VTON decomposes the virtual try-on task into three stages, incorporates a human parsing model is to guide the image synthesis, a Warp-GAN learns to synthesize the realistic image by alleviating misalignment caused by diverse pose, and a refinement render recovers the texture details. We construct a new dataset for the multi-pose guided virtual try-on task covering person images with more poses and clothes diversity. Extensive experiments demonstrate that our MG-VTON significantly outperforms all state-of-the-art methods both qualitatively and quantitatively with promising performances.

{\small
\bibliographystyle{ieee}
\bibliography{egbib}

\begin{thebibliography}{10}\itemsep=-1pt

\bibitem{balakrishnan2018synthesizing}
G.~Balakrishnan, A.~Zhao, A.~V. Dalca, F.~Durand, and J.~Guttag.
\newblock Synthesizing images of humans in unseen poses.
\newblock In {\em CVPR}, 2018.

\bibitem{belongie2002shape}
S.~Belongie, J.~Malik, and J.~Puzicha.
\newblock Shape matching and object recognition using shape contexts.
\newblock {\em IEEE TPAMI}, 24(4):509--522, 2002.

\bibitem{bookstein1989tps}
F.~L. Bookstein.
\newblock Principal warps: Thin-plate splines and the decomposition of
  deformations.
\newblock {\em IEEE TPAMI}, 11(6):567--585, 1989.

\bibitem{cao2017openpose}
Z.~Cao, T.~Simon, S.-E. Wei, and Y.~Sheikh.
\newblock Realtime multi-person 2d pose estimation using part affinity fields.
\newblock In {\em CVPR}, 2017.

\bibitem{Esser2018vunet}
P.~Esser, E.~Sutter, and B.~Ommer.
\newblock A variational u-net for conditional appearance and shape generation.
\newblock In {\em CVPR}, 2018.

\bibitem{gong2017look}
K.~Gong, X.~Liang, X.~Shen, and L.~Lin.
\newblock Look into person: Self-supervised structure-sensitive learning and a
  new benchmark for human parsing.
\newblock In {\em CVPR}, 2017.

\bibitem{goodfellow2014generative}
I.~Goodfellow, J.~Pouget-Abadie, M.~Mirza, B.~Xu, D.~Warde-Farley, S.~Ozair,
  A.~Courville, and Y.~Bengio.
\newblock Generative adversarial nets.
\newblock In {\em NIPS}, 2014.

\bibitem{han2017viton}
X.~Han, Z.~Wu, Z.~Wu, R.~Yu, and L.~S. Davis.
\newblock Viton: An image-based virtual try-on network.
\newblock In {\em CVPR}, 2018.

\bibitem{isola2017pix2pix}
P.~Isola, J.-Y. Zhu, T.~Zhou, and A.~A. Efros.
\newblock Image-to-image translation with conditional adversarial networks.
\newblock In {\em CVPR}, 2017.

\bibitem{jetchev2017conditional}
N.~Jetchev and U.~Bergmann.
\newblock The conditional analogy gan: Swapping fashion articles on people
  images.
\newblock {\em ICCVW}, 2(6):8, 2017.

\bibitem{johnson2016perceptual}
J.~Johnson, A.~Alahi, and L.~Fei-Fei.
\newblock Perceptual losses for real-time style transfer and super-resolution.
\newblock In {\em ECCV}, pages 694--711, 2016.

\bibitem{kim2017discoGAN}
T.~Kim, M.~Cha, H.~Kim, J.~Lee, and J.~Kim.
\newblock Learning to discover cross-domain relations with generative
  adversarial networks.
\newblock {\em arXiv preprint arXiv:1703.05192}, 2017.

\bibitem{Diederik2014Adam}
D.~P. Kingma and J.~Ba.
\newblock Adam: {A} method for stochastic optimization.
\newblock {\em arXiv preprint arXiv:1412.6980}, 2014.

\bibitem{laehner2018deepwrinkles}
Z.~Laehner, D.~Cremers, and T.~Tung.
\newblock Deepwrinkles: Accurate and realistic clothing modeling.
\newblock In {\em ECCV}, 2018.

\bibitem{lassner2017generative}
C.~Lassner, G.~Pons-Moll, and P.~V. Gehler.
\newblock A generative model of people in clothing.
\newblock In {\em CVPR}, 2017.

\bibitem{ledig2016photo}
C.~Ledig, L.~Theis, F.~Husz{\'a}r, J.~Caballero, A.~Cunningham, A.~Acosta,
  A.~Aitken, A.~Tejani, J.~Totz, Z.~Wang, et~al.
\newblock Photo-realistic single image super-resolution using a generative
  adversarial network.
\newblock In {\em CVPR}, 2017.

\bibitem{ma2017pose}
L.~Ma, X.~Jia, Q.~Sun, B.~Schiele, T.~Tuytelaars, and L.~Van~Gool.
\newblock Pose guided person image generation.
\newblock In {\em NIPS}, 2017.

\bibitem{ma2017disentangled}
L.~Ma, Q.~Sun, S.~Georgoulis, L.~Van~Gool, B.~Schiele, and M.~Fritz.
\newblock Disentangled person image generation.
\newblock In {\em CVPR}, 2018.

\bibitem{mirza2014cgan}
M.~Mirza and S.~Osindero.
\newblock Conditional generative adversarial nets.
\newblock {\em arXiv preprint arXiv:1411.1784}, 2014.

\bibitem{pons2017clothcap}
G.~Pons-Moll, S.~Pujades, S.~Hu, and M.~J. Black.
\newblock Clothcap: Seamless 4d clothing capture and retargeting.
\newblock {\em ACM Transactions on Graphics (TOG)}, 36(4):73, 2017.

\bibitem{pumarola2018unsupervised}
A.~Pumarola, A.~Agudo, A.~Sanfeliu, and F.~Moreno-Noguer.
\newblock Unsupervised person image synthesis in arbitrary poses.
\newblock In {\em CVPR}, 2018.

\bibitem{reed2016text2image}
S.~Reed, Z.~Akata, X.~Yan, L.~Logeswaran, B.~Schiele, and H.~Lee.
\newblock Generative adversarial text-to-image synthesis.
\newblock In {\em ICML}, 2016.

\bibitem{Rocco2017geocnn}
I.~Rocco, R.~Arandjelovi\'c, and J.~Sivic.
\newblock Convolutional neural network architecture for geometric matching.
\newblock In {\em CVPR}, 2017.

\bibitem{ronn2015unet}
O.~Ronneberger, P.~Fischer, and T.~Brox.
\newblock U-net: Convolutional networks for biomedical image segmentation.
\newblock In {\em MICCAI}, pages 234--241, 2015.

\bibitem{NIPS2016_6125}
T.~Salimans, I.~Goodfellow, W.~Zaremba, V.~Cheung, A.~Radford, X.~Chen, and
  X.~Chen.
\newblock Improved techniques for training gans.
\newblock In {\em NIPS}, 2016.

\bibitem{sekine2014virtual}
M.~Sekine, K.~Sugita, F.~Perbet, B.~Stenger, and M.~Nishiyama.
\newblock Virtual fitting by single-shot body shape estimation.
\newblock In {\em International Conference on 3d Body Scanning Technologies},
  pages 406--413, 2014.

\bibitem{siarohin2017deformable}
A.~Siarohin, E.~Sangineto, S.~Lathuiliere, and N.~Sebe.
\newblock Deformable gans for pose-based human image generation.
\newblock {\em arXiv preprint arXiv:1801.00055}, 2017.

\bibitem{simonyan2015very}
K.~Simonyan and A.~Zisserman.
\newblock Very deep convolutional networks for large-scale image recognition.
\newblock In {\em ICLR}, 2015.

\bibitem{wang2018cpvton}
B.~Wang, H.~Zhang, X.~Liang, Y.~Chen, and L.~Lin.
\newblock Toward characteristic-preserving image-based virtual try-on network.
\newblock In {\em ECCV}, 2018.

\bibitem{wang2017pix2pixHD}
T.-C. Wang, M.-Y. Liu, J.-Y. Zhu, A.~Tao, J.~Kautz, and B.~Catanzaro.
\newblock High-resolution image synthesis and semantic manipulation with
  conditional gans.
\newblock In {\em CVPR}, 2018.

\bibitem{wang2004image}
Z.~Wang, A.~C. Bovik, H.~R. Sheikh, and E.~P. Simoncelli.
\newblock Image quality assessment: from error visibility to structural
  similarity.
\newblock {\em TIP}, 13(4):600--612, 2004.

\bibitem{yan2017skeleton}
Y.~Yan, J.~Xu, B.~Ni, W.~Zhang, and X.~Yang.
\newblock Skeleton-aided articulated motion generation.
\newblock In {\em ACM MM}, 2017.

\bibitem{Yang2017inpainting}
C.~Yang, X.~Lu, Z.~Lin, E.~Shechtman, O.~Wang, and H.~Li.
\newblock High-resolution image inpainting using multi-scale neural patch
  synthesis.
\newblock In {\em CVPR}, 2017.

\bibitem{yi2017dualgan}
Z.~Yi, H.~Zhang, P.~Tan, and M.~Gong.
\newblock Dualgan: Unsupervised dual learning for image-to-image translation.
\newblock {\em arXiv preprint}, 2017.

\bibitem{zhang2017detailed}
C.~Zhang, S.~Pujades, M.~J. Black, and G.~Pons-Moll.
\newblock Detailed, accurate, human shape estimation from clothed 3d scan
  sequences.
\newblock In {\em CVPR}, volume~2, page~3, 2017.

\bibitem{zhu2017cycleGAN}
J.-Y. Zhu, T.~Park, P.~Isola, and A.~A. Efros.
\newblock Unpaired image-to-image translation using cycle-consistent
  adversarial networks.
\newblock In {\em ICCV}, 2017.

\bibitem{zhu2017fashionGAN}
S.~Zhu, S.~Fidler, R.~Urtasun, D.~Lin, and C.~C. Loy.
\newblock Be your own prada: Fashion synthesis with structural coherence.
\newblock In {\em ICCV}, 2017.

\bibitem{liu2016deepfashion}
S.~Q. X.~W. Ziwei~Liu, Ping~Luo and X.~Tang.
\newblock Deepfashion: Powering robust clothes recognition and retrieval with
  rich annotations.
\newblock In {\em CVPR}, pages 1096--1104, 2016.

\end{thebibliography}
}




\begin{figure*}[!h]
\centering
\includegraphics[width=1.0\hsize \hspace{0.01\hsize}]{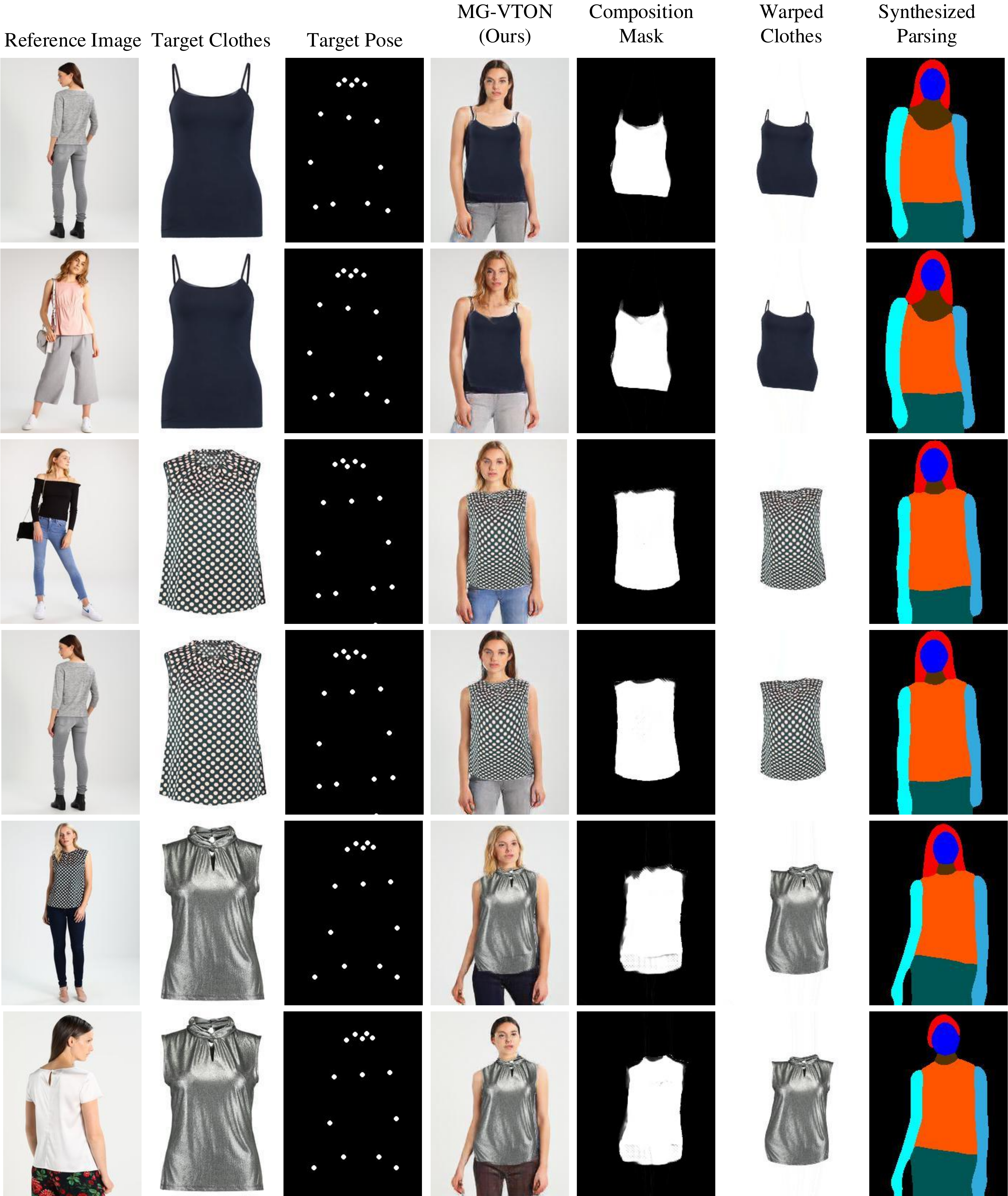} 
\caption{Test results of our MG-VTON on MPV dataset.}
\label{fig:ours2}
\end{figure*}

\begin{figure*}[!h]
\centering
\includegraphics[width=1.0\hsize \hspace{0.01\hsize}]{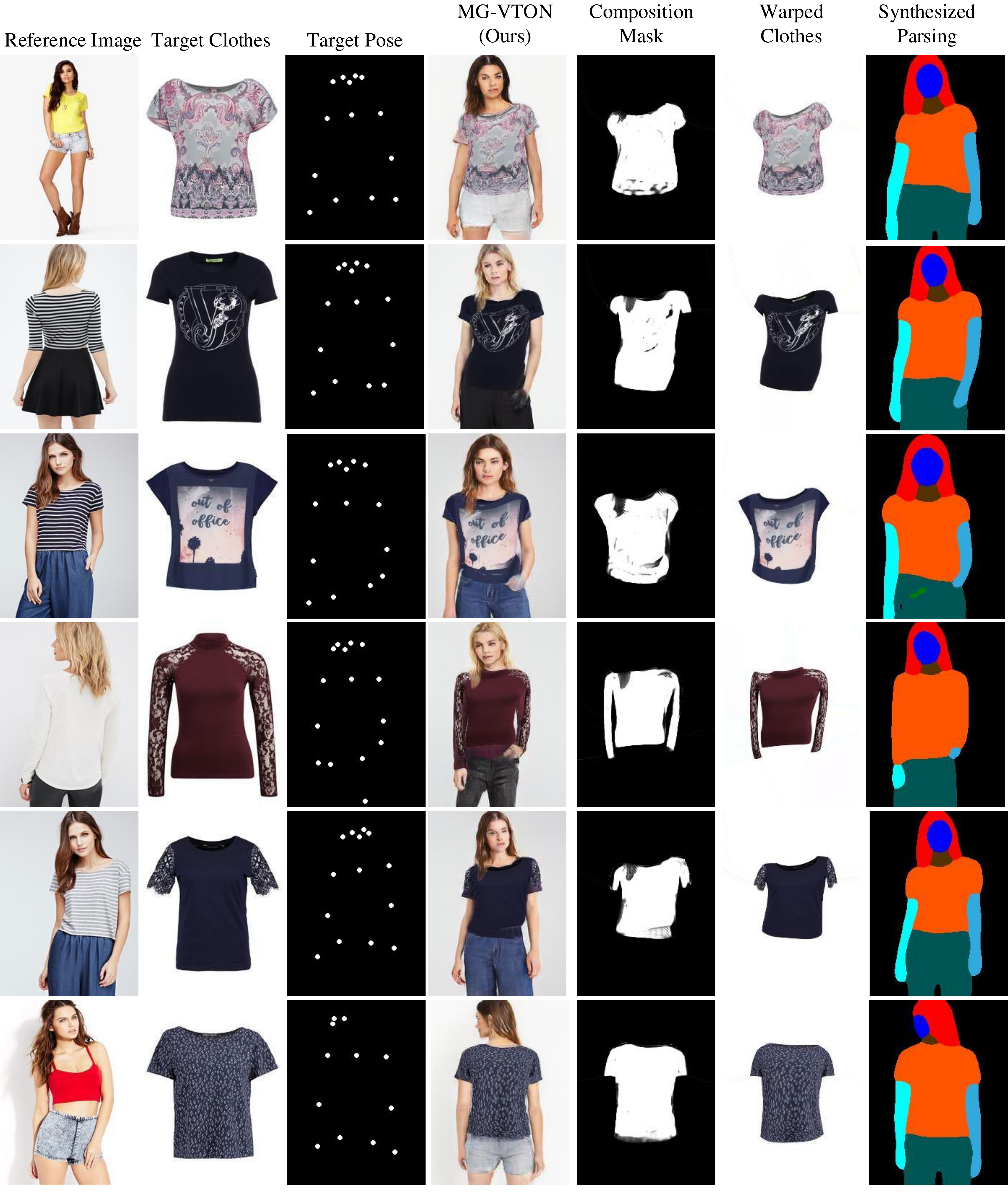} 
\caption{Test results of our MG-VTON, trained on MPV dataset, test on DeepFashion dataset.}
\label{fig:ours3}
\end{figure*}

\end{document}